\pdfoutput=1

\documentclass[11pt]{article}
\usepackage{enumitem}

\usepackage[preprint]{acl}

\usepackage{times}
\usepackage{url}
\usepackage{latexsym}

\usepackage[T1]{fontenc}

\usepackage[utf8]{inputenc}

\usepackage{microtype}

\usepackage{inconsolata}
\usepackage{inconsolata}
\usepackage{multirow}
\usepackage{makecell}
\usepackage{booktabs}
\usepackage{pifont}
\usepackage{bm}
\usepackage{xspace,mfirstuc,tabulary}
\usepackage{amsmath,amsfonts,amssymb,amstext,bm}
\usepackage{mathrsfs}
\usepackage{mathrsfs}
\usepackage{colortbl}
\usepackage{graphicx}  
\usepackage{float}  
\usepackage{subfigure} 
\usepackage{soul}
\usepackage{dcolumn} 
\usepackage{inconsolata}
\usepackage{hyperref}
\usepackage{siunitx}
\usepackage{enumitem}
\usepackage{caption}
\usepackage{subfigure}
\usepackage{caption}
\usepackage{graphicx}
\usepackage{float} 
\usepackage{subfigure}
\usepackage{microtype}
\usepackage{tikz-dependency}
\usepackage[linguistics]{forest}
\usepackage{caption}
\usepackage{subcaption}
\usepackage{enumitem}
\usepackage{soul}
\usepackage{tikz-qtree}
\usepackage[T1]{fontenc}
\usepackage[utf8]{inputenc}
\usepackage{CJKutf8}
\usepackage{CJK}
\usepackage{utfsym}
\usepackage{booktabs}
\usepackage{tabularx}
\usepackage{tcolorbox}
\usepackage{subfiles}
\usepackage{todonotes}
\usepackage{marginnote}
\usepackage{paralist}
\usetikzlibrary{fit, calc, decorations.pathreplacing, positioning, shapes.geometric, backgrounds}
\usepackage{graphicx}

\newcommand\wz{\phantom{0}}

\title{
Self-Correction Makes LLMs Better Parsers \\}

\author{Ziyan Zhang, Yang Hou, Chen Gong\thanks{\; Corresponding author.}, Zhenghua Li \\  
 School of Computer Science and Technology, Soochow University \\
 \texttt{\{zyzhang0509,yhou1\}@stu.suda.edu.cn} \\
 \texttt{\{gongchen18,zhli13\}@suda.edu.cn}
 }

\begin{document}
\begin{CJK*}{UTF8}{gkai}

\maketitle
\begin{abstract}

Large language models (LLMs) have achieved remarkable success across various natural language processing (NLP) tasks. However, recent studies suggest that they still face challenges in performing fundamental NLP tasks essential for deep language understanding, particularly syntactic parsing.
In this paper, we conduct an in-depth analysis of LLM parsing capabilities, delving into the specific shortcomings of their parsing results.
We find that LLMs may stem from limitations to fully leverage grammar rules in existing treebanks, which restricts their capability to generate valid syntactic structures.
To help LLMs acquire knowledge without additional training, we propose a self-correction method that leverages grammar rules from existing treebanks to guide LLMs in correcting previous errors.
Specifically, we automatically detect potential errors and dynamically search for relevant rules, offering hints and examples to guide LLMs in making corrections themselves. 
Experimental results on three datasets with various LLMs, demonstrate that our method significantly improves performance in both in-domain and cross-domain settings {on the English and Chinese datasets}.

\end{abstract}

\section{Introduction}

LLMs have exhibited significant success across a wide range of NLP tasks and applications \citep{NEURIPS2022_8bb0d291, raunak-etal-2023-leveraging, chiang-lee-2023-large}.
However, in the field of syntax parsing, enabling LLMs to achieve a deep structural understanding of sentences remains challenging \citep{bai2023constituencyparsingusingllms,li-etal-2023-llm,tian-etal-2024-large}.
Although non-LLM parsers have achieved over 95\% F-scores in the newswire domain, the performance of LLMs falls significantly short.
This discrepancy inevitably raises two questions: 
Where exactly does the parsing capability of LLMs fall short? What specific knowledge are LLMs lacking? 

Prior works have provided some findings about the parsing capabilities of LLMs.
For instance, \citet{bai2023constituencyparsingusingllms} pointed out hallucinations in LLM outputs, while \citet{tian-etal-2024-large} highlighted that they tend to be much flatter than gold trees.
However, existing works primarily analyze high-level performance metrics without delving into the detailed shortcomings of LLM parsing or the underlying causes.
To address these gaps, we conduct an in-depth analysis of LLM parsing capabilities, examining both structural characteristics of their outputs and the specific types of errors they exhibit.
Our findings suggest that the inability to fully utilize grammar rules in the training data is a critical limitation of LLMs.
Unlike non-LLM parsers, which can comprehensively learn rules from extensive training data, LLMs solely rely on limited contextual examples provided during few-shot learning.
This fundamental difference distinguishes LLM parsing performance from non-LLM parsers.

Motivated by the above analysis, we are interested in methods that allow LLMs to
learn structural knowledge without additional training.
In this work, we propose a self-correction method that first identifies errors in the initial answer, then searches the treebank for relevant examples, and finally guides LLMs in correcting errors via hints and examples.
This method effectively enables LLMs to reduce all types of errors in the parsing results, thereby improving the overall parsing performance.
Experimental results prove that our method achieves significant improvement across different LLMs in both in-domain and cross-domain settings {on the English and Chinese datasets}.

In conclusion, our contributions are three-fold:
\begin{asparaitem}[$\bullet$]
    \item We conduct an in-depth analysis of LLM parsing capability from the perspective of the overall parsing performance, the characteristics of their parsing results over grammar rules, and the types of errors they made.
    \item We propose a self-correction method that enable LLMs to learn from existing treebanks and correct all types of errors in their previous predictions, enhancing the parsing capability without the need for additional training.
    \item We conduct extensive experiments on various LLMs, and our method achieves consistent improvements across different datasets. Further analysis demonstrates that our method effectively mitigates the limitation of overly flatter parsing results from LLMs.
\end{asparaitem}
\section{In-depth Analysis of LLM Parsing Capability}
LLMs have demonstrated impressive performance in a wide range of NLP tasks \citep{NEURIPS2020_1457c0d6, NEURIPS2022_9d560961}. 
However, recent studies indicate that LLMs face significant challenges 
when performing syntactic parsing tasks, e.g., 
generating invalid bracket format strings, or 
producing parse trees that are flatter than gold trees 
\citep{bai2023constituencyparsingusingllms,tian-etal-2024-large}.

Although previous studies have explored the parsing capability of LLMs to some extent, their analysis remains relatively limited in providing deep insights into the limitations of LLMs, as both studies lack a comprehensive analysis of the reason behind the weak performance and the specific error types made by LLMs. 
To gain a deeper understanding of the constituency parsing capacity of LLMs, we conduct an in-depth analysis from perspectives of the overall parsing performance, the characteristics of their parsing results, and the types of errors they made.

Specifically, first, we evaluate the parsing performance of different LLMs under the few-shot setting, which is compared to one of the non-LLM SOTA models, the Berkeley parser \citep{kitaev-klein-2018-constituency}, to investigate the overall parsing capabilities of existing LLMs.
Second, we compare the parsing results of LLMs with both gold trees and that from SOTA non-LLM parsers, aiming to analyze the distinct characteristics of LLMs and non-LLM parsers, as well as reasons for the significant performance gap between them.
Finally, we categorize the parsing errors made by LLMs into four types and analyze the error distribution, conducting an in-depth examination to reveal which types of errors LLMs commonly make.

\subsection{Overall Parsing Performance of LLMs}\label{sec:baseline}

\begin{table}[t]
\centering
\small
\begin{tabular}{l|l|cccccc}
\toprule
\textbf{Dataset}& \textbf{Model} & \textbf{R}& \textbf{P}& \textbf{F}   \\
\midrule
\multirow{5}{*}{\makecell{PTB}} &Berkeley& 95.56&  96.10 & 95.82  \\
    &LLaMa-8B& 11.39&   11.91 &11.64 \\
   &LLaMa-70B & 45.02& 51.24 & 47.92 \\
 &GPT-3.5  & 62.24 &71.62 & 66.60 \\
&GPT-4  & 69.97 &  77.14 &  73.38   \\
\midrule
\multirow{5}{*}{CTB5} &Berkeley& 88.76&   92.82 & 90.74  \\
&LLaMa-8B& 7.01&  11.68 & 8.77   \\
&Qwen-72B& 34.45&  45.60 & 39.24   \\
&DeeSeekp-v3& 43.49& 52.71 & 47.65  \\
 &GPT-3.5  & 27.19 &45.84 & 34.14 \\
 
&GPT-4  & 40.44 &  49.95 &  44.69   \\
\midrule
 \multirow{5}{*}{MCTB} &Berkeley  &86.38 &88.50 & 87.43\\
 &LLaMa-8B& 9.67&  9.48 & 9.57 \\
 &LLaMa-70B   & 40.77&48.40 & 44.26 \\
 &GPT-3.5  & 51.71 &62.71 & 56.68 \\
 
&GPT-4  & 62.29 & 67.98 & 65.01    \\
\bottomrule
\end{tabular}
\caption{The overall parsing performance of different models on the PTB, CTB5, and MCTB datasets.}
\label{table:baseline}
\vspace{-0.9em}
\end{table}

We evaluate the capability of LLMs in both in-domain and cross-domain settings. 
For in-domain evaluation, we use the original test sets of the Penn Treebank (PTB) \citep{marcus-etal-1993-building} and Chinese Penn Treebank 5 (CTB5) \citep{Xue2005ThePC}.
For cross-domain evaluation, to manage costs, we use a subset of the Multi-domain Constituent Treebank (MCTB) \citep{yang-etal-2022-challenges}, randomly sampling 200 sentences from each domain of MCTB, which covers Dialogue, Forum, Law, Literature, and Review domains, resulting in a 1,000 sentences test set. {We examine whether these datasets have been leaked to the LLMs in Appendix \ref{sec:MKP}}.
Our experiments include both open-sourced LLMs, i.e., LLaMA-3-8B
\footnote{We use the English version of LLaMA-3 from \href{https://huggingface.co/meta-llama}{https://huggingface.co/meta-llama} and the Chinese version from \href{https://github.com/ymcui/Chinese-LLaMA-Alpaca-3}{https://github.com/ymcui/Chinese-LLaMA-Alpaca-3}.}, LLaMA-3-70B \citep{llama3}, Qwen-2.5-72B \cite{qwen2} and DeeSeekp-v3 \cite{deepseekai2024deepseekv3technicalreport}, and closed-sourced LLMs, i.e., GPT-3.5 \footnote{\href{https://platform.openai.com/docs/models/gpt-3.5-turbo-1106}{https://platform.openai.com/docs/models/gpt-3.5-turbo}} and GPT-4 \footnote{\href{https://platform.openai.com/docs/models/gpt-4}{https://platform.openai.com/docs/models/gpt-4}}.
To understand the gap between LLMs and non-LLMs, we also report the performance of Berkeley Neural Parser \citep{kitaev-klein-2018-constituency}, an existing SOTA neural parser. 
For the in-domain setting, we train Berkeley on the train set of PTB and CTB5, respectively. For the cross-domain setting, we train Berkeley on the train set of PTB and evaluate on the MCTB.

Following standard practice in previous works, we sample five examples in PTB/CTB-train for the few-shot setting. After generation, we process the tree valid by adding left or right brackets when the brackets of LLMs are mismatched.
Finally, we evaluate the performance of models with the standard evaluation toolkit EVALB\footnote{\href{https://nlp.cs.nyu.edu/evalb/}{https://nlp.cs.nyu.edu/evalb/}}.
Details of the LLM parsing prompt are shown in Appendix \ref{sec:baseline_prompt}.

Table \ref{table:baseline} presents the results of various LLMs on different test sets. 
From the results, we can draw the following conclusions. First, the performance of all LLMs shows a significant decrease compared to that of the traditional non-LLM parser. This may be because the non-LLM parser has thoroughly learned the structures present in the existing treebanks while LLMs may not. We will discuss this in more detail in Section \ref{sec:known_rule}. Second, among all LLMs, closed-source LLMs (GPT-3.5 and GPT-4) deliver better results than open-source LLMs. 
Third, when adapted to the cross-domain test set, all the parsers exhibit a large performance drop. This indicates a notable gap between specific-domain and general-domain data, highlighting the greater challenges posed by cross-domain parsing.

The above results naturally give rise to two important considerations: 1) the specific limitations of LLMs in parsing capabilities compared to non-LLM parsers, and 2) the unique characteristics of LLM-generated parse results in comparison to gold-standard parse trees.
Therefore, we conduct a more in-depth analysis in Section \ref{sec:known_rule} and Section \ref{sec:error_type}.

\subsection{Analysis over Grammar Rules}\label{sec:known_rule}

Previous study \citep{dakota-kubler-2021-whats} has proved that parsers tend to generate structures that have been seen in the train set, and most of these known structures are parsed correctly. 
\footnote{In this paper, ``structures'' refers to rules, which are defined as subtrees invlove a parent node and its child nodes. For example, the subtree ``the proposed \$ 7 billion bill'' in Figure \ref{fig:label_error} corresponds to the rule ``NP $\rightarrow$ DT VBN ADJP''. }
Unlike non-LLM parsers that can learn from the entire train set, LLMs are limited to a few examples in the prompt, thus lacking approaches to all the grammar rules.
To investigate whether these known grammar rules in the train set are significant factors that affect the performance, we present rule statistics of the parsing results from different models in Table \ref{tab:known_rule}. 
The third column represents the absolute number of rules in the parsing trees. 
The fourth column gives the number of known rules in the parsed results. 
The fifth and sixth columns show the accuracy of known rules and unknown rules in the parsing results, respectively.

\begin{table}[ht]
    \centering\small
    \setlength{\tabcolsep}{3pt}
    \begin{tabular}{l l l l l l}
        \toprule
        \multirow{2}{*}{\textbf{Dataset}}& \multirow{2}{*}{\textbf{Model}} & \multicolumn{2}{c}{\textbf{\# Total }}& \multicolumn{2}{c}{\textbf{Accuracy (\%)}} \\
        & & parsed& known& known & unknown  \\
        \midrule
        \multirow{5}{*}{\makecell{PTB}} &Berkeley&   46,441 & 26,745 & 96.20 & 60.18 \\
        &LLaMa-8B &   42,710 & 13,616 & 26.88&  \wz1.93\\
           &LLaMa-70B & 38,895 & 19,597 & 68.11 & 19.82\\
         &GPT-3.5   &38,958& 18,409 & 79.47 & 33.11 \\
         
        &GPT-4   & 40,166  &  20,291 & 81.44 & 37.14 \\
        \midrule
        \multirow{5}{*}{CTB5} &Berkeley&  \wz8,691 & \wz6,198 & 90.19&  76.82\\
        &LLaMa-8B & \wz5,792& \wz1,993 & 36.78 & \wz4.53\\
        &Qwen-72B& \wz6,851& \wz3,043 & 59.19 & 13.81 \\
        &DeeSeekp-v3& \wz7,491& \wz4,259 & 64.94  & 23.33\\
         &GPT-3.5   & \wz5,480 & \wz2,489 & 55.85 & 11.07\\
         
        &GPT-4   &  \wz7,342 &  \wz3,479  &  59.73 &  17.19   \\
        \midrule
         \multirow{5}{*}{MCTB} &Berkeley&  17,744 & \wz9,916 & 86.24 & 68.19  \\
         &LLaMa-8B &   17,587 & \wz5,688 & 35.07 & \wz2.15 \\
         &LLaMa-70B&   14,457 & \wz7,518 & 64.34 &   16.50  \\
         &GPT-3.5   &14,155 & \wz7,091 & 73.01& 27.34\\   
        &GPT-4   & 15,731 & \wz8,329 & 73.54& 33.27 \\
        \bottomrule
    \end{tabular}
    \caption{Rule statistics of the parsing results from different models across different datasets.}
    \label{tab:known_rule}
\end{table}

We observe that the parsing results from non-LLM parsers contain more numbers of known rules compared to those from LLMs.
Further analyzing the accuracy of these rules, we find that known rules from non-LLM achieves a high accuracy rate of 83.32\% to 95.76\%
, whereas those from LLMs are much lower.
This suggests that non-LLM parsers effectively learn common and valid rules from existing treebanks and apply them during parsing on different domains. 
For LLMs, performance improves as the total number and accuracy of known rules increase. 
Therefore, we speculate that the total number and accuracy of known rules are the main factors affecting the performance, and the known rules are the specific knowledge that LLMs lack.
The significant performance gap between LLMs and non-LLM parsers may arise from LLMs not systematically learning these rules during training, which increases the likelihood of generating invalid structures during parsing.

\subsection{Analysis over Four Types of Errors}\label{sec:error_type}

\begin{figure}[tb]
\centering
\flushleft \footnotesize \emph{Predicted Subtree} $\qquad$$\qquad$$\qquad$$\qquad$ $\qquad$ \emph{Gold Subtree}
\subfigure[Span Error ]
{
\label{fig:span_error}
  \begin{minipage}[b]{1.0\columnwidth}
      \centering
      \begin{tikzpicture}[
          scale=0.7,
          level distance=17pt,
          every tree node/.style={align=center,anchor=base},
          frontier/.style={distance from root=50pt},
          postag/.style={rounded corners=1mm,align=center},
          error/.style={fill={rgb,255:red,221; green,68; blue,80}, draw=red!80, thick, fill opacity=0.2, text opacity=1, rounded corners=0.1mm,align=center},
          edge from parent/.style={draw,edge from parent path={(\tikzparentnode.south) {[rounded corners=0.1pt]-- ($(\tikzchildnode |- \tikzparentnode.south) + (0, -5pt)$) -- (\tikzchildnode)}}}
        ]
        \begin{scope}[xshift=-50pt]
            \Tree
            [.{NP} 
                [.{DT} {a} ]
                [.{JJ} {young} ]
                [.{NN} {man} ]
              ];           
        \end{scope}
        \begin{scope}[xshift=100pt]
            \Tree
            [.{NP} 
                [.{NP}
                [.{DT} {a} ]
                [.{JJ} {young} ]
                [.{NN} {man} ]
                [.{POS} {'s} ]
                ]
                [.{NN} {sport} ]
                ]];    
        \end{scope}
        \draw[blue,dashed] (0.7,-0.9) rectangle(3.3,-1.9) node[pos=.5] {};
      \end{tikzpicture}\\
  \end{minipage}
}
\flushleft \footnotesize 
\emph{Predicted Subtree} $\qquad$$\qquad$$\qquad$$\qquad$ $\qquad$ \emph{Gold Subtree}
\subfigure[Label Error]
{
\label{fig:label_error}
    \begin{minipage}[b]{1.0\columnwidth}
      \centering
      \begin{tikzpicture}[
          scale=0.7,
          level distance=21pt,
          every tree node/.style={align=center,anchor=base},
          frontier/.style={distance from root=45pt},
          postag/.style={rounded corners=1mm,align=center},
          error/.style={fill={rgb,255:red,221; green,68; blue,80}, draw=red!80, thick, fill opacity=0.2, text opacity=1, rounded corners=0.01mm,align=center},
          edge from parent/.style={draw,edge from parent path={(\tikzparentnode.south) {[rounded corners=0.1pt]-- ($(\tikzchildnode |- \tikzparentnode.south) + (0, -5pt)$) -- (\tikzchildnode)}}}
        ]
        \begin{scope}[xshift=-80pt]
            \Tree
            [.{NP} 
                [.\colorbox [RGB]{255,228,240}{DT} 
                [.{the}
                ]]
                [.\colorbox [RGB]{255,228,240}{VBN} 
                [.{proposed} 
                ]]
                [.\colorbox [RGB]{255,228,240}{QP}
                [.{\$ 7 billion bill}
                ]
                ]];       
        \end{scope}
        \begin{scope}[xshift=65pt]
            \Tree
            [.{NP} 
                [.\colorbox [RGB]{255,228,240}{DT} 
                [.{the}
                ]]
                [.\colorbox [RGB]{255,228,240}{VBN} 
                [.{proposed} 
                ]]
                [.\colorbox [RGB]{255,228,240}{ADJP}
                [.{\$ 7 billion bill}
                ]
                ]];        
        \end{scope}
      \end{tikzpicture}\\
\flushleft \scriptsize  \emph{Label:}$\quad $\tiny \colorbox [RGB]{255,242,221}{DT\quad VBN\quad QP} $\qquad$$\qquad$$\qquad$$\qquad$ \colorbox [RGB]{255,242,221}{DT\quad VBN\quad ADJP}
  \end{minipage}
}\\
\flushleft \footnotesize \emph{Predicted Subtree} $\qquad$$\qquad$$\qquad$$\qquad$ $\qquad$ \emph{Gold Subtree}
\subfigure[Flatness Error ]
{
\label{fig:flatness_error}
  \begin{minipage}[b]{1.0\columnwidth}
      \centering
      \begin{tikzpicture}[
          scale=0.7,
          level distance=21pt,
          every tree node/.style={align=center,anchor=base},
          frontier/.style={distance from root=40pt},
          postag/.style={rounded corners=1mm,align=center},
          error/.style={fill={rgb,255:red,221; green,68; blue,80}, draw=red!80, thick, fill opacity=0.2, text opacity=1, rounded corners=0.01mm,align=center},
          edge from parent/.style={draw,edge from parent path={(\tikzparentnode.south) {[rounded corners=0.1pt]-- ($(\tikzchildnode |- \tikzparentnode.south) + (0, -6pt)$) -- (\tikzchildnode)}}}
        ]
        \begin{scope}[xshift=-110pt]
            \Tree
            [.{NP} 
                [.\colorbox [RGB]{255,228,240}{JJ} {senior} ]
                [.\colorbox [RGB]{255,228,240}{NN} {vice} ]
                [.\colorbox [RGB]{255,228,240}{NN} {president} ]
                [.\colorbox [RGB]{255,228,240}{PP} {at Crop}
              ]];      
        \end{scope}
        \begin{scope}[xshift=60pt]
            \Tree
            [.{NP} 
                [.{NP} 
                [.\colorbox [RGB]{255,228,240}{JJ} 
                [.{senior}
                ]]
                [.\colorbox [RGB]{255,228,240}{NN} 
                [.{vice} 
                ]]
                [.\colorbox [RGB]{255,228,240}{NN}
                [.{president}
                ]
                ]]
                [.\colorbox [RGB]{255,228,240}{PP}
                [.{at Crop}
                ]]];    
        \end{scope}
      \end{tikzpicture}\\
\flushleft \scriptsize  \emph{Label:}$\quad $\tiny\colorbox [RGB]{255,242,221}{JJ \quad NN\quad NN\quad PP } $\qquad$$\qquad$ $\qquad$ $\qquad$  \tiny \colorbox [RGB]{255,242,221}{JJ \quad NN\quad NN\quad PP }
  \end{minipage}
}\\
\flushleft \footnotesize \emph{Predicted Subtree} $\qquad$$\qquad$$\qquad$$\qquad$ $\qquad$ \emph{Gold Subtree}
\subfigure[Deepness Error]
{
\label{fig:deepness_error}
  \begin{minipage}[b]{1.0\columnwidth}
      \centering
      \begin{tikzpicture}[
          scale=0.7,
          level distance=20pt,
          every tree node/.style={align=center,anchor=base},
          frontier/.style={distance from root=45pt},
          postag/.style={rounded corners=1mm,align=center},
          error/.style={fill={rgb,255:red,221; green,68; blue,80}, draw=red!80, thick, fill opacity=0.2, text opacity=1, rounded corners=0.01mm,align=center},
          edge from parent/.style={draw,edge from parent path={(\tikzparentnode.south) {[rounded corners=0.1pt]-- ($(\tikzchildnode |- \tikzparentnode.south) + (0, -5pt)$) -- (\tikzchildnode)}}}
        ]
        \begin{scope}[xshift=-75pt]
            \Tree
            [.{S} 
                [.\colorbox [RGB]{255,228,240}{CC} 
                [.{But} ]]
                [.{S}
                [.\colorbox [RGB]{255,228,240}{NP}
                [.{students}
                ]]
                [.\colorbox [RGB]{255,228,240}{VP}
                [.{laugh loudly}
                ]]]]];        
        \end{scope}
        \begin{scope}[xshift=75pt]
            \Tree
            [.{S} 
                [.\colorbox [RGB]{250,228,240}{CC} 
                [.{But}
                ]]
                [.\colorbox [RGB]{255,228,240}{NP}
                [.{students}
                ]]
                [.\colorbox [RGB]{255,228,240}{NP}
                [.{laugh loudly}
                ]]];    
        \end{scope}
      \end{tikzpicture}\\
\flushleft \scriptsize  \emph{Label:}$\quad $\tiny\colorbox [RGB]{255,242,221}{ CC\quad NP\quad VP} $\qquad$$\qquad$ $\qquad$ $\qquad$  $\qquad$ \tiny \colorbox [RGB]{255,242,221}{ CC\quad NP\quad NP}
\vspace{0.01em}
  \end{minipage}
}\\

\caption{Four types of errors.}
\label{fig:four_error}
\end{figure}
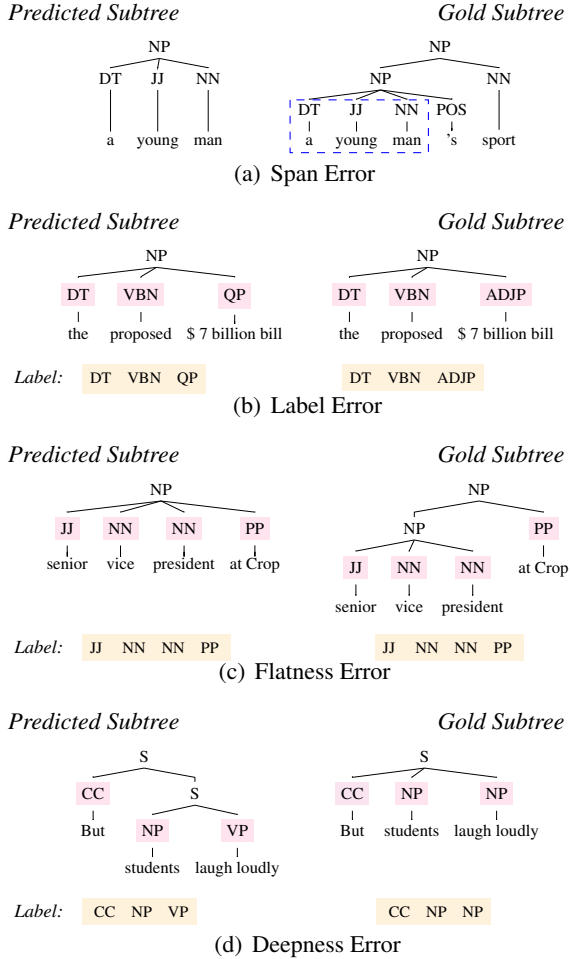

\begin{figure}[ht]
    \centering
    \includegraphics[width=\linewidth]{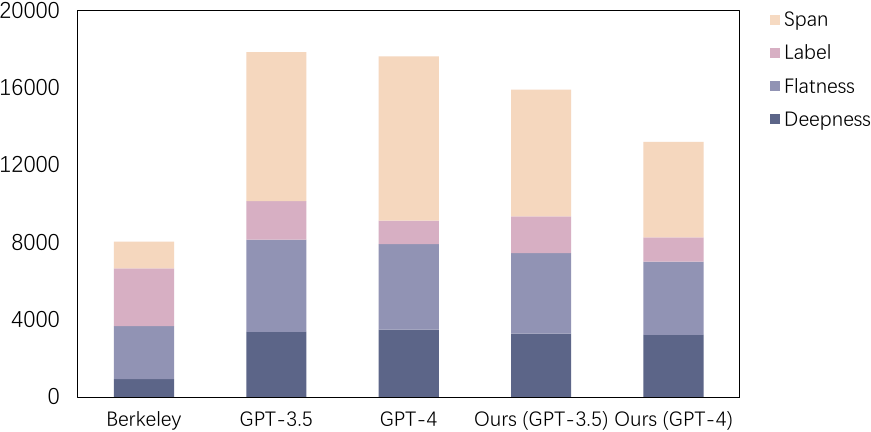}
    \caption{The overview of the four types of errors made by different models on the PTB.}
    \label{fig:error_type}
\end{figure}

Motivated by previous works \cite{dakota-kubler-2021-whats,kummerfeld-etal-2013-empirical}, we divide parsing errors into four types, as illustrated in 
Figure \ref{fig:four_error}. 
Note that we define parsing errors of subtrees by considering two levels: the root node and its child nodes, while disregarding the internal structure of the lower layers within the subtree.

\paragraph{Span Errors.}
We define ``span'' as a sequence of words from $word_i$ to $word_j$ that forms a constituent and we do not consider the label, as it will be involved when identifying label errors.
Therefore, span error represents that the boundary of the parent node does not align with the gold. For instance, the span of the subtree ``a young man'' in Figure \ref{fig:span_error} ranging from ``a'' to ``man'' does not exist in the gold ``a young man 's sport''.

\paragraph{Label Errors.}
The predicted subtree has the same span
with the gold subtree, but one or more constituent labels are predicted incorrectly, as shown in Figure \ref{fig:label_error}. It could be an error in the label of the parent node or the child nodes. 

\paragraph{Flatness Errors.}
The predicted subtree is flatter than the gold subtree, which represents the parent node in the predicted subtree contains more child nodes, as shown in Figure \ref{fig:flatness_error}.

\paragraph{Deepness Errors.}
Contrary to flatness errors, the predicted subtree is deeper than the gold subtree, which represents the parent node contains fewer child nodes, as shown in Figure \ref{fig:deepness_error}.

Given a subtree, we first determine whether it is a span error or one of the other three error types by checking if the span of the parent node 
corresponding to the subtree appears in the gold tree. Furthermore, we use the number of child nodes to determine which of the three error types it is. 
In this way, these four types of errors can encompass all the mistakes made by the parsers.
Certainly, compound situations may occur, such as a label error within a flatness error or a deepness error, but we do not further categorize these.

Figure \ref{fig:error_type} shows an overview of the error types made by Berkeley, GPT-3.5 and GPT-4 on the PTB. 
As expected, the total number of errors increases as the performance of models declines, and LLMs make significantly more errors than non-LLMs.
For flatness error, it accounts for a relatively large proportion across all the models.  
For deepness error, the parsing results of LLM contain fewer deepness errors than those of non-LLM in the in-domain setting. 
We hypothesize that this is due to the characteristic of LLMs to generate flatter structures, which results in a greater number of flatness errors. 
Because of this characteristic, there are fewer instances where the predicted trees are deeper than the gold trees, leading to a fewer number of deepness errors compared to non-LLMs in the in-domain setting.
{We provide the distributions of errors across different models and different dataset in Appendix \ref{sec:all_errors}}.

\section{Our Self-Correction Approach}\label{sec:self_correct}

\begin{figure*}[ht]
    \centering
    \includegraphics[width=\linewidth]{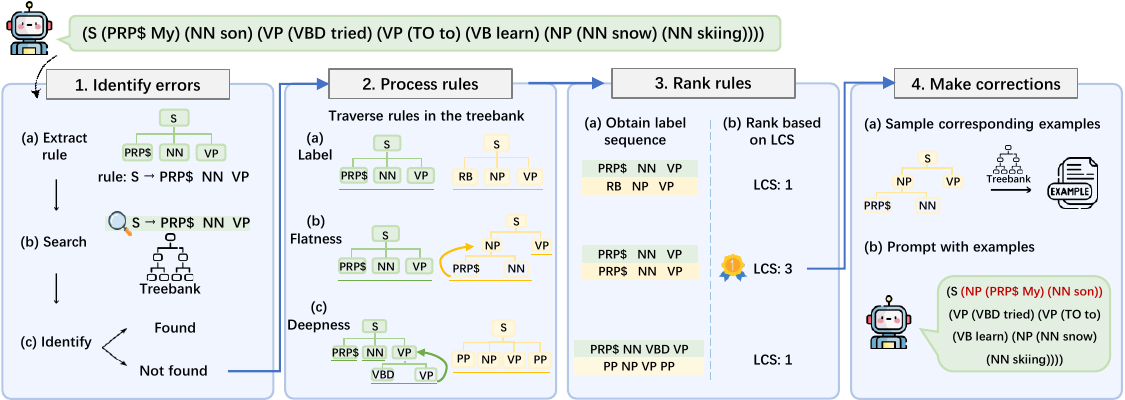}
    \caption{The process of structure correction. For clarify, we provide a typical example of rule for each processing method in the figure. In practice, each rule needs to undergo three types of processing methods and be ranked.}
    \label{fig:structure_correction}
\end{figure*}
The above results and analysis help us better understand the shortcomings of LLM parsing results and encourage us to propose targeted methods for enhancing LLM parsing capabilities.
Since Section \ref{sec:known_rule} has proved that the low performance of LLMs may stem from their lack of known rules in the existing treebanks, 
thereby we are interested in exploring methodologies that enable LLMs to effectively assimilate knowledge from the existing treebanks without the need for additional training, thereby improving their constituency parsing capabilities. 
Furthermore, we aim at a method based on the four types of parsing errors mentioned in Section \ref{sec:error_type} to effectively address all errors made by the LLM.
In this work, we propose a self-correction method based on the rules from the existing treebank, guiding LLMs to correct errors in the previous answers.
Consistent with the common process of self-correction in previous works \citep{pan-etal-2024-automatically, tong-etal-2024-llms}, we first prompt LLM to generate a base answer, identify potential errors in it, and then guide LLM to correct these errors through specific hints and examples.

Specifically, based on the characteristics of the parsing results, we categorize the self-correction method into two parts: unmatch correction and structure correction.
First, we perform unmatch correction to ensure the alignment of the leaf nodes between the predicted trees and gold trees.
Then we make structure correction, which addresses the four types of structural errors introduced in \ref{sec:error_type}.

\subsection{Unmatch Correction}
We find that LLMs often make changes on their own, leading to parse trees containing new words which do not exist in the original sentence. This phenomenon has also been demonstrated by \citet{bai2023constituencyparsingusingllms}, leading to a significant impact on the performance. Therefore, we first make unmatch correction. After obtaining base answers, we identify errors and design different hints for different errors, letting LLM generate a new answer. The detail of hints is shown in Appendix
\ref{sec:unmatch_correct}.

\subsection{Structure Correction}
After ensuring the alignment of the leaf nodes between predicted trees and gold trees, we make structure correction.
Specifically, we identify errors by comparing them with grammar rules in the existing treebank and correct subtrees of different heights from the top to the bottom.
Compared with previous works that focus on the design of the prompt, the advantages of our method are three-fold:
1) Considering that the number and accuracy of known rules are decisive factors influencing performance, we leverage existing treebanks to guide LLM parsing in a training-free manner, enabling LLMs to quickly acquire structural knowledge.
2) Our method preserves the correct parts of the parsing results and focuses solely on correcting the potentially incorrect parts, which greatly saves time and resources.
3) Our correction covers all types of parsing errors. For different errors, our proposed method can dynamically search rules in the existing treebanks to serve as hints to guide the correction.
As is shown in Figure \ref{fig:structure_correction}, our structure correction can be divided into four detailed steps:

\textbf{1) Identifying Errors Based on Rules.}
To identify errors in the base answer, we extract rules of the subtree and search it in the existing treebanks\footnote{We use the train sets of PTB and CTB5 as the existing treebanks for English and Chinese, respectively.} .
If the rule is not found in the treebank, we determine it is likely incorrect and requires correction.

\textbf{2) Processing Rules based on Errors.}
Once an error is identified, we aim at searching rules that can guide the correction of the current error.
Therefore, we traverse rules in the existing treebanks and process them based on the characteristics of different errors so that the correction involves all types of errors. 
The main idea is that if the predicted rule and the traversed rule can be transformed into the same form after applying a specific processing method, we can infer that the predicted rule has made the corresponding error.
As is shown in Figure \ref{fig:four_error}, for label error, we directly take the predicted rule and the traversed rule.
For flatness error, we replace the child nodes of the traversed rule with their own child nodes.
For deepness error, on the contrary to flatness error, we replace the child nodes of the predicted rule with their own child nodes.

\textbf{3) Ranking Based on the Similarity.} 
After processing, our goal is to find rules most similar to the gold trees to serve as examples. 
If two rules exhibit significant similarity following the application of specific processing method, 
the traversed rules are considered close to the gold, and can provide valuable structural information to LLMs.
To determine the similarity, we take the constituent labels of the child nodes to form label sequences as shown in Figure \ref{fig:structure_correction} and use the Longest Common Subsequence (LCS) as the metric.
{If there are rules with the same LCS, we prioritize the rule that appears more frequently in the treebank.}\label{line:714-716}
Then we rank traversed rules and select the top five rules, considering them can guide the LLM in self-corrections.

\textbf{4) Guiding LLM in Making Self-Corrections.}
Finally, we sample examples corresponding to these rules from the treebank, adding them to the prompt and guiding LLM to generate a new answer.

Through different processing methods, our method can directly correct label , flatness, and deepness error.
Different from these three types, span error occurs outside the subtree (rules) and cannot be resolved by making corrections to the subtrees.
However, since our method starts from the top of the entire tree and correct subtrees of different heights from top to bottom,
we speculate that this process can also indirectly help resolve span errors, which has been proven in Experiments \ref{sec:analysis_error_types}. 
Additionally, we only consider the constituency labels within the subtree, ignoring the label of root node. This is because we assume that the label of root node has already been corrected during the correction of higher subtrees. 
Also, setting a constraint that the label of root node must match the predicted rules during traversing significantly helps reduce time.
{We also investigate the effect of different numbers of the final rules and different searching strategies in Appendix \ref{sec:setting} and \ref{sec:pos}}.
\section{Experiments}
 
 

\begin{table}[t]
\centering
\small
\resizebox{\columnwidth}{!}{
\begin{tabular}{l|l|cccccc}
\toprule
\textbf{Dataset}& \textbf{Model} & \textbf{R}& \textbf{P}& \textbf{F}   \\
\midrule
\multirow{6}{*}{\makecell{PTB}} 
 &GPT-3.5$^\dagger$  &72.48 & 84.86 & 78.18 \\
  &GPT-4$^\dagger$  & 77.74 & \textbf{89.04} & 83.00 \\
  \cmidrule{2-5}
   &Llama-8B& 29.42\textsubscript{+18.03} & 38.27\textsubscript{+26.36} & 33.27\textsubscript{+21.63}  \\
   &Llama-70B & 54.90\textsubscript{+9.88} & 61.49\textsubscript{+10.25} & 58.01\textsubscript{+10.09}  \\
 &GPT-3.5  & 74.59\textsubscript{+12.35} & 80.11\textsubscript{\wz+8.49} & 77.25\textsubscript{+10.65} \\
&GPT-4  & \textbf{82.27}\textsubscript{+12.30} & 84.78\textsubscript{\wz+7.64}  &  \textbf{83.50}\textsubscript{+10.12}  \\
\midrule
\multirow{4}{*}{CTB5} &Llama-8B& 22.63\textsubscript{+15.62} & 34.22\textsubscript{+22.54} & 27.24\textsubscript{+18.47}  \\
&Qwen-72B& 40.12\textsubscript{\wz+5.67} & 52.18\textsubscript{\wz+6.58} & 45.36\textsubscript{\wz+6.12}  \\
&DeeSeekp-v3& 57.33\textsubscript{+13.84} & 65.08\textsubscript{+12.37} & 60.96\textsubscript{+13.31} \\
 &GPT-3.5  & 39.50\textsubscript{+12.31} & 57.12\textsubscript{+11.28} & 46.70\textsubscript{+12.56} \\
 
&GPT-4  & \textbf{61.67}\textsubscript{+21.23} & \textbf{68.69}\textsubscript{+18.74} &  \textbf{64.99}\textsubscript{+20.30}   \\
\midrule
 \multirow{4}{*}{MCTB} &Llama-8B  & 25.91\textsubscript{+16.24} & 34.01\textsubscript{+24.53} & 29.41\textsubscript{+19.84} \\
 &Llama-70B & 48.31\textsubscript{\wz+7.54} & 55.68\textsubscript{\wz+7.28} & 51.73\textsubscript{\wz+7.47} \\
 &GPT-3.5  & 60.14\textsubscript{\wz+8.43} & 71.71\textsubscript{\wz+8.99} & 65.41\textsubscript{\wz+8.73}  \\
 
&GPT-4  & \textbf{71.71}\textsubscript{\wz+9.42} & \textbf{77.13}\textsubscript{\wz+9.15} & \textbf{74.32}\textsubscript{\wz+9.31}  \\
\bottomrule
\end{tabular}
}
\caption{The main result of our self-correction method. The "\textbf{Bold}" identifies the best performance. The $\dagger$ denotes the results referred from \citet{tian-etal-2024-large}. The subscript \textsubscript{+} indicates the improvement.}
\label{table:main_result}
\vspace{-0.9em}
\end{table}

We conduct experiments to verify the effectiveness of our self-correction method and perform a detailed analysis to gain insights into what LLMs have learned through our method.
All experimental settings remain consistent with those in direct LLM-based parsing, as described in Section \ref{sec:baseline}.

\subsection{Main Results}
The main experimental results of our method are listed in Table~\ref{table:main_result}.
Compared to the results of
directly using LLM for parsing with five-shot learning in Table \ref{table:baseline}
, our proposed self-correction method achieves substantial and consistent improvements across all datasets. 
In particular, for the in-domain setting, our method achieves an improvement of over 10 F1 scores on both PTB and CTB5.
For cross-domain settings, despite the gap between the general treebank and the target domain test set, our method still brings consistent improvements. 
These results demonstrate that our method enables LLMs to learn structural knowledge applicable across different domains from existing treebanks, thus comprehensively improving the parsing ability.

Compared to the result of \citet{tian-etal-2024-large}, which first prompt LLMs for chunking and then use the chunking results to guide LLM parsing, though our result with GPT-3.5 is 0.93 F1 score lower on PTB, the result with GPT-4 is 0.5 F1 score higher. 
We hypothesize that this is because our method requires LLMs to perform self-correction by learning from examples and making independent judgments by themselves. Therefore, the stronger the capability of an LLM, the greater its performance improves.
Moreover, the improvements in recall are particularly significant, indicating that our method encourages LLMs to parse deeper trees with more rules. This effectively mitigates a major weakness of LLMs in constituency parsing, i.e., the tendency to produce overly flat parse trees than gold trees.

\textit{Overall, our method consistently achieves significant improvements under both in-domain and cross-domain settings , effectively mitigating the limitation of overly flatter parsing results from LLMs.}

\begin{figure}[t!]
    \centering
    \includegraphics[width=\linewidth]{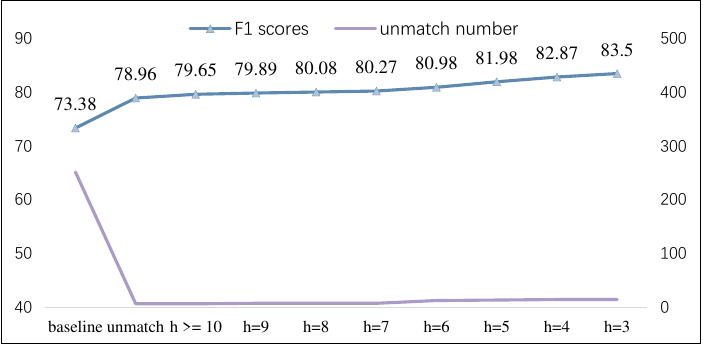}
    \caption{The effect of unmatch correction and structure correction with GPT-4 on the PTB dataset. ``unmatch'' represents unmatch correction and ``h'' represents the height of subtrees in the structure correction.}
    \label{fig:different_correction}
\end{figure}
\subsection{The Effect of Different Corrections}

To analyze the effectiveness of unmatch correction and structure correction, we examine the improvements brought by unmatch correction and structure correction at different tree heights with GPT-4 on the PTB.
As is shown in Figure \ref{fig:different_correction}, first, LLMs generate a substantial number of unmatched trees during parsing, which greatly affects the parsing performance. 
By implementing our method, the number of unmatched trees is significantly reduced, leading to a marked improvement in the performance.
Second, structure correction also significantly enhances the performance. Specifically, the improvement rate of the performance increases as the height of subtrees decrease, which means correcting lower subtrees is more effective. 
This may be because the rules of lower subtrees have a higher overlap rate between the treebank and the test set.

\begin{figure}[ht]
    \centering
    \includegraphics[width=\linewidth]{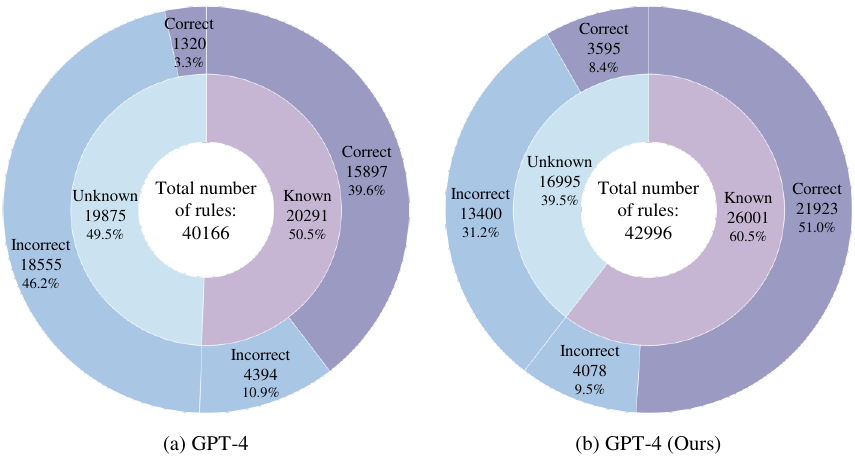}
    \caption{Rule statistics of the parsing results generated from GPT-4 before and after applying our self-correction method on the PTB dataset.}
    \label{fig:known_rules_our_method_v1}
\end{figure}
\subsection{Has LLMs Truly Acquired Knowledge?}

To investigate whether the LLM truly learns structural knowledge during self-correction, we compare the rule statistic before and after applying our method with GPT-4 on the PTB.
As shown in Figure \ref{fig:known_rules_our_method_v1}, we categorize rules into known rules and unknown rules and further analyze their accuracy, respectively.
First, our method significantly increases the absolute number of rules, bringing it closer to that of the gold trees. 
Second, consistent with our motivation that
enable LLMs to assimilate knowledge from existing treebank without training
as illustrated in Section \ref{sec:self_correct}, our method effectively enhances both the number of known rules in parsing results and their accuracy. 
Surprisingly, our method also raises the accuracy of unknown rules.
This demonstrates that our method not only enables LLMs to quickly learn knowledge from the treebanks without training, but also enhances their capability to generate correct rules, regardless of whether they have encountered those rules before.
{We also conducte experiment on whether our method add new errors in Appendix \ref{sec:add}}.

\subsection{Which Types Do LLMs Correct?}\label{sec:analysis_error_types}
Although our method aims at correcting four types of parsing errors, we remain curious about which specific error types the LLMs actually correct. 
Therefore, we analyze the number of different errors after applying our method on the GPT-3.5 and GPT-4.  
As is shown in Figure \ref{fig:error_type}, the number of all types of errors reduces, verifying the effectiveness of our method. 
Our method addresses span errors the most, which are the most frequent errors in the parsing results. As we hypothesized, although span errors cannot be directly corrected, our method can indirectly reduce these errors through the top-down correction process.
The number of the other three error types also reduce significantly, 
demonstrating that our method enables LLMs to identify and prioritize rules similar to the gold ones.

\subsection{The Effect of Our Method on the Unknown Rules}\label{sec:unknown}
{Our method aims to enhance the parsing performance of LLMs via known rules, but we are also curious whether it can contributes to predicting unknown rules. And our experimental results show that after applying our method, the accuracy of unknown rules increased from 35.3\% to 48.0\%. This demonstrates that our method not only does not limit the capability of LLMs to handle unknown rules but also comprehensively improves the parsing capability through known rules, resulting in a significant increase in accuracy for unknownn rules. This might be because, although some rules do not completely align with those in the treebank, LLMs can learn similar structural information or some local information from the examples, which also helps improve the their parsing capability.}

\begin{table}[t]
    \centering
    \begin{tabular}{c ccc}
        \toprule        
        Method &   \textbf{R}   &   \textbf{P} &   \textbf{F}\\
        \midrule    
        Random selection  &  71.47  &  79.25  &  75.16 \\  
        Our method  &  \textbf{82.27}  &  \textbf{84.78}  &  \textbf{83.50} \\ 
        \bottomrule
    \end{tabular}
    \caption{Results of random selection and our method on the PTB with GPT-4.}
    \label{tab:random_selection}
\end{table}  
\subsection{Random Selection vs. Our Method}\label{sec:random_selection}

{To further validate the effectiveness of our method, we also compared it with randomly selecting examples for few-shoting prompting on the PTB with GPT-4.
As indicated in Table \ref{tab:random_selection}, our results significantly surpass that of random selection,  demonstrating that our method can select more suitable examples for LLMs via ranking based on LCS.}
\section{Related Work}

\paragraph{LLM Parsing.} 
LLMs have achieved remarkable success in various tasks and applications \citep{NEURIPS2022_8bb0d291,NEURIPS2022_9d560961}. 
However, recent studies indicate that LLMs exhibit week capability in parsing and show a noticeable gap compared to that of non-LLM parsers.
\citet{bai2023constituencyparsingusingllms} revealed that LLMs suffer from hallucinations and have limited ability to learn extremely long constituents. 
\citet{zhou2023largelanguagemodelsunderstand} conducted experiments on 24 LLMs and found that most of them have a limited grasp of syntactic knowledge.
\citet{tian-etal-2024-large} observed that LLMs are shallow parsers in that they are ineffective at conducting fullparsing. They propose a three-step approach that firstly use LLMs for chunking and then add the chunks to guide LLMs for parsing.
These previous works have gained insights into the parsing capability of LLMs. However, they lack deeper analysis of the reasons behind the low performance and the characteristics of their parsing results.
We first conduct an in-depth analysis of LLM parsing capability, investigating the detailed shortcomings of their parsing results and the underlying causes.
Based on these analyses, we design a method that enables LLMs to obtain knowledge from the existing treebank, enhancing their parsing ability significantly.

\paragraph{Self-Correction.}
To unleash the reasoning abilities of LLMs, many recent work focus on design prompt to instruct LLMs to solve problems with human-like logic, such as Chain-of-Thought \citep{NEURIPS2022_9d560961}, Self-Consistency \citep{wang2023selfconsistencyimproveschainthought} and Self-Correction \citep{gao2023rarr, madaan2024self}.
Self-Correction, which is defined as the process of continuously evaluating and adjusting previous responses to get better answers, has attracted widespread researches due to its alignment with typical human learning strategy and its notable effectiveness across various NLP tasks.
It can be divided into two types: one is that LLMs correct entirely on their own knowledge and ability; The other is for LLMs to correct with external feedback, such as knowledge sources \citet{gao-etal-2023-rarr}, other models or tools \citep{yang-etal-2022-re3}.
Recent studies have questioned the efficiency of intrinsic Self-Correction of LLMs.
\citet{huang2023large} revealed that without external feedback, it is difficult for LLMs to correct their own mistakes.
\citet{tyen2023llms} proved that LLMs mainly fail in finding errors in the answers, but can correct them given the error location.
To tackle these challenges, we propose a self-correction method that automatically identifies errors in the parsing results and leverages the existing treebanks as external feedback to guide the corrections, thereby enhancing the parsing capability of LLMs.

\section{Conclusion}
In this paper, we conduct an in-depth analysis of LLM parsing capability from the perspective of the overall parsing performance, 
the characteristics of their parsing results based on the rules, and 
the types of errors they made. 
Based on the analysis that the low performance of LLMs may stem from their lack of known rules in existing treebanks, we propose a self-correction method that enables LLMs to efficiently acquire rules from treebanks without additional training, thereby correcting errors in the previous answer.
Experimental results demonstrate that our method consistently improves performance, effectively guiding LLMs to correct all types of errors made by themselves.

\section{Limitations}
There are several unexplored avenues of interest, such as determining whether rules similar but not identical to the gold trees can guide LLMs in making corrections, and assessing the extent to which LLMs actually learn from examples. These will be addressed in future studies.
\bibliography{custom}

\begin{thebibliography}{28}
\providecommand{\natexlab}[1]{#1}

\bibitem[{Bai et~al.(2023)Bai, Wu, Chen, Wang, and Zhang}]{bai2023constituencyparsingusingllms}
Xuefeng Bai, Jialong Wu, Yulong Chen, Zhongqing Wang, and Yue Zhang. 2023.
\newblock \href {https://arxiv.org/abs/2310.19462} {Constituency parsing using llms}.
\newblock \emph{Preprint}, arXiv:2310.19462.

\bibitem[{Brown et~al.(2020)Brown, Mann, Ryder, Subbiah, Kaplan, Dhariwal, Neelakantan, Shyam, Sastry, Askell, Agarwal, Herbert-Voss, Krueger, Henighan, Child, Ramesh, Ziegler, Wu, Winter, Hesse, Chen, Sigler, Litwin, Gray, Chess, Clark, Berner, McCandlish, Radford, Sutskever, and Amodei}]{NEURIPS2020_1457c0d6}
Tom Brown, Benjamin Mann, Nick Ryder, Melanie Subbiah, Jared~D Kaplan, Prafulla Dhariwal, Arvind Neelakantan, Pranav Shyam, Girish Sastry, Amanda Askell, Sandhini Agarwal, Ariel Herbert-Voss, Gretchen Krueger, Tom Henighan, Rewon Child, Aditya Ramesh, Daniel Ziegler, Jeffrey Wu, Clemens Winter, Chris Hesse, Mark Chen, Eric Sigler, Mateusz Litwin, Scott Gray, Benjamin Chess, Jack Clark, Christopher Berner, Sam McCandlish, Alec Radford, Ilya Sutskever, and Dario Amodei. 2020.
\newblock \href {https://proceedings.neurips.cc/paper_files/paper/2020/file/1457c0d6bfcb4967418bfb8ac142f64a-Paper.pdf} {Language models are few-shot learners}.
\newblock In \emph{Advances in Neural Information Processing Systems}, volume~33, pages 1877--1901. Curran Associates, Inc.

\bibitem[{Chiang and Lee(2023)}]{chiang-lee-2023-large}
Cheng-Han Chiang and Hung-yi Lee. 2023.
\newblock \href {https://doi.org/10.18653/v1/2023.acl-long.870} {Can large language models be an alternative to human evaluations?}
\newblock In \emph{Proceedings of the 61st Annual Meeting of the Association for Computational Linguistics (Volume 1: Long Papers)}, pages 15607--15631, Toronto, Canada. Association for Computational Linguistics.

\bibitem[{Dakota and K{\"u}bler(2021)}]{dakota-kubler-2021-whats}
Daniel Dakota and Sandra K{\"u}bler. 2021.
\newblock \href {https://aclanthology.org/2021.scil-1.29} {What{'}s in a span? evaluating the creativity of a span-based neural constituency parser}.
\newblock In \emph{Proceedings of the Society for Computation in Linguistics 2021}, pages 323--333, Online. Association for Computational Linguistics.

\bibitem[{DeepSeek-AI et~al.(2024)DeepSeek-AI, Liu, Feng, Xue, Wang, Wu, Lu, Zhao, Deng, Zhang, Ruan, Dai, Guo, Yang, Chen, Ji, Li, Lin, Dai, Luo, Hao, Chen, Li, Zhang, Bao, Xu, Wang, Zhang, Ding, Xin, Gao, Li, Qu, Cai, Liang, Guo, Ni, Li, Wang, Chen, Chen, Yuan, Qiu, Li, Song, Dong, Hu, Gao, Guan, Huang, Yu, Wang, Zhang, Xu, Xia, Zhao, Wang, Zhang, Li, Wang, Zhang, Zhang, Tang, Li, Tian, Huang, Wang, Zhang, Wang, Zhu, Chen, Du, Chen, Jin, Ge, Zhang, Pan, Wang, Xu, Zhang, Chen, Li, Lu, Zhou, Chen, Wu, Ye, Ye, Ma, Wang, Zhou, Yu, Zhou, Pan, Wang, Yun, Pei, Sun, Xiao, Zeng, Zhao, An, Liu, Liang, Gao, Yu, Zhang, Li, Jin, Wang, Bi, Liu, Wang, Shen, Chen, Zhang, Chen, Nie, Sun, Wang, Cheng, Liu, Xie, Liu, Yu, Song, Shan, Zhou, Yang, Li, Su, Lin, Li, Wang, Wei, Zhu, Zhang, Xu, Xu, Huang, Li, Zhao, Sun, Li, Wang, Yu, Zheng, Zhang, Shi, Xiong, He, Tang, Piao, Wang, Tan, Ma, Liu, Guo, Wu, Ou, Zhu, Wang, Gong, Zou, He, Zha, Xiong, Ma, Yan, Luo, You, Liu, Zhou, Wu, Ren, Ren, Sha, Fu, Xu, Huang, Zhang, Xie, Zhang, Hao,
  Gou, Ma, Yan, Shao, Xu, Wu, Zhang, Li, Gu, Zhu, Liu, Li, Xie, Song, Gao, and Pan}]{deepseekai2024deepseekv3technicalreport}
DeepSeek-AI, Aixin Liu, Bei Feng, Bing Xue, Bingxuan Wang, Bochao Wu, Chengda Lu, Chenggang Zhao, Chengqi Deng, Chenyu Zhang, Chong Ruan, Damai Dai, Daya Guo, Dejian Yang, Deli Chen, Dongjie Ji, Erhang Li, Fangyun Lin, Fucong Dai, Fuli Luo, Guangbo Hao, Guanting Chen, Guowei Li, H.~Zhang, Han Bao, Hanwei Xu, Haocheng Wang, Haowei Zhang, Honghui Ding, Huajian Xin, Huazuo Gao, Hui Li, Hui Qu, J.~L. Cai, Jian Liang, Jianzhong Guo, Jiaqi Ni, Jiashi Li, Jiawei Wang, Jin Chen, Jingchang Chen, Jingyang Yuan, Junjie Qiu, Junlong Li, Junxiao Song, Kai Dong, Kai Hu, Kaige Gao, Kang Guan, Kexin Huang, Kuai Yu, Lean Wang, Lecong Zhang, Lei Xu, Leyi Xia, Liang Zhao, Litong Wang, Liyue Zhang, Meng Li, Miaojun Wang, Mingchuan Zhang, Minghua Zhang, Minghui Tang, Mingming Li, Ning Tian, Panpan Huang, Peiyi Wang, Peng Zhang, Qiancheng Wang, Qihao Zhu, Qinyu Chen, Qiushi Du, R.~J. Chen, R.~L. Jin, Ruiqi Ge, Ruisong Zhang, Ruizhe Pan, Runji Wang, Runxin Xu, Ruoyu Zhang, Ruyi Chen, S.~S. Li, Shanghao Lu, Shangyan Zhou, Shanhuang
  Chen, Shaoqing Wu, Shengfeng Ye, Shengfeng Ye, Shirong Ma, Shiyu Wang, Shuang Zhou, Shuiping Yu, Shunfeng Zhou, Shuting Pan, T.~Wang, Tao Yun, Tian Pei, Tianyu Sun, W.~L. Xiao, Wangding Zeng, Wanjia Zhao, Wei An, Wen Liu, Wenfeng Liang, Wenjun Gao, Wenqin Yu, Wentao Zhang, X.~Q. Li, Xiangyue Jin, Xianzu Wang, Xiao Bi, Xiaodong Liu, Xiaohan Wang, Xiaojin Shen, Xiaokang Chen, Xiaokang Zhang, Xiaosha Chen, Xiaotao Nie, Xiaowen Sun, Xiaoxiang Wang, Xin Cheng, Xin Liu, Xin Xie, Xingchao Liu, Xingkai Yu, Xinnan Song, Xinxia Shan, Xinyi Zhou, Xinyu Yang, Xinyuan Li, Xuecheng Su, Xuheng Lin, Y.~K. Li, Y.~Q. Wang, Y.~X. Wei, Y.~X. Zhu, Yang Zhang, Yanhong Xu, Yanhong Xu, Yanping Huang, Yao Li, Yao Zhao, Yaofeng Sun, Yaohui Li, Yaohui Wang, Yi~Yu, Yi~Zheng, Yichao Zhang, Yifan Shi, Yiliang Xiong, Ying He, Ying Tang, Yishi Piao, Yisong Wang, Yixuan Tan, Yiyang Ma, Yiyuan Liu, Yongqiang Guo, Yu~Wu, Yuan Ou, Yuchen Zhu, Yuduan Wang, Yue Gong, Yuheng Zou, Yujia He, Yukun Zha, Yunfan Xiong, Yunxian Ma, Yuting Yan, Yuxiang
  Luo, Yuxiang You, Yuxuan Liu, Yuyang Zhou, Z.~F. Wu, Z.~Z. Ren, Zehui Ren, Zhangli Sha, Zhe Fu, Zhean Xu, Zhen Huang, Zhen Zhang, Zhenda Xie, Zhengyan Zhang, Zhewen Hao, Zhibin Gou, Zhicheng Ma, Zhigang Yan, Zhihong Shao, Zhipeng Xu, Zhiyu Wu, Zhongyu Zhang, Zhuoshu Li, Zihui Gu, Zijia Zhu, Zijun Liu, Zilin Li, Ziwei Xie, Ziyang Song, Ziyi Gao, and Zizheng Pan. 2024.
\newblock \href {https://arxiv.org/abs/2412.19437} {Deepseek-v3 technical report}.
\newblock \emph{Preprint}, arXiv:2412.19437.

\bibitem[{Gao et~al.(2023{\natexlab{a}})Gao, Dai, Pasupat, Chen, Chaganty, Fan, Zhao, Lao, Lee, Juan, and Guu}]{gao-etal-2023-rarr}
Luyu Gao, Zhuyun Dai, Panupong Pasupat, Anthony Chen, Arun~Tejasvi Chaganty, Yicheng Fan, Vincent Zhao, Ni~Lao, Hongrae Lee, Da-Cheng Juan, and Kelvin Guu. 2023{\natexlab{a}}.
\newblock \href {https://doi.org/10.18653/v1/2023.acl-long.910} {{RARR}: Researching and revising what language models say, using language models}.
\newblock In \emph{Proceedings of the 61st Annual Meeting of the Association for Computational Linguistics (Volume 1: Long Papers)}, pages 16477--16508, Toronto, Canada. Association for Computational Linguistics.

\bibitem[{Gao et~al.(2023{\natexlab{b}})Gao, Dai, Pasupat, Chen, Chaganty, Fan, Zhao, Lao, Lee, Juan et~al.}]{gao2023rarr}
Luyu Gao, Zhuyun Dai, Panupong Pasupat, Anthony Chen, Arun~Tejasvi Chaganty, Yicheng Fan, Vincent Zhao, Ni~Lao, Hongrae Lee, Da-Cheng Juan, et~al. 2023{\natexlab{b}}.
\newblock Rarr: Researching and revising what language models say, using language models.
\newblock In \emph{Proceedings of the 61st Annual Meeting of the Association for Computational Linguistics (Volume 1: Long Papers)}, pages 16477--16508.

\bibitem[{Huang et~al.(2023)Huang, Chen, Mishra, Zheng, Yu, Song, and Zhou}]{huang2023large}
Jie Huang, Xinyun Chen, Swaroop Mishra, Huaixiu~Steven Zheng, Adams~Wei Yu, Xinying Song, and Denny Zhou. 2023.
\newblock Large language models cannot self-correct reasoning yet.
\newblock \emph{arXiv preprint arXiv:2310.01798}.

\bibitem[{Kitaev and Klein(2018)}]{kitaev-klein-2018-constituency}
Nikita Kitaev and Dan Klein. 2018.
\newblock \href {https://doi.org/10.18653/v1/P18-1249} {Constituency parsing with a self-attentive encoder}.
\newblock In \emph{Proceedings of the 56th Annual Meeting of the Association for Computational Linguistics (Volume 1: Long Papers)}, pages 2676--2686, Melbourne, Australia. Association for Computational Linguistics.

\bibitem[{Kojima et~al.(2022)Kojima, Gu, Reid, Matsuo, and Iwasawa}]{NEURIPS2022_8bb0d291}
Takeshi Kojima, Shixiang~(Shane) Gu, Machel Reid, Yutaka Matsuo, and Yusuke Iwasawa. 2022.
\newblock \href {https://proceedings.neurips.cc/paper_files/paper/2022/file/8bb0d291acd4acf06ef112099c16f326-Paper-Conference.pdf} {Large language models are zero-shot reasoners}.
\newblock In \emph{Advances in Neural Information Processing Systems}, volume~35, pages 22199--22213. Curran Associates, Inc.

\bibitem[{Kummerfeld et~al.(2013)Kummerfeld, Tse, Curran, and Klein}]{kummerfeld-etal-2013-empirical}
Jonathan~K. Kummerfeld, Daniel Tse, James~R. Curran, and Dan Klein. 2013.
\newblock \href {https://aclanthology.org/P13-2018/} {An empirical examination of challenges in {C}hinese parsing}.
\newblock In \emph{Proceedings of the 51st Annual Meeting of the Association for Computational Linguistics (Volume 2: Short Papers)}, pages 98--103, Sofia, Bulgaria. Association for Computational Linguistics.

\bibitem[{Li et~al.(2023)Li, Zhang, Guo, Zhang, and Zhang}]{li-etal-2023-llm}
Jianling Li, Meishan Zhang, Peiming Guo, Min Zhang, and Yue Zhang. 2023.
\newblock \href {https://doi.org/10.18653/v1/2023.emnlp-main.508} {{LLM}-enhanced self-training for cross-domain constituency parsing}.
\newblock In \emph{Proceedings of the 2023 Conference on Empirical Methods in Natural Language Processing}, pages 8174--8185, Singapore. Association for Computational Linguistics.

\bibitem[{Madaan et~al.(2024)Madaan, Tandon, Gupta, Hallinan, Gao, Wiegreffe, Alon, Dziri, Prabhumoye, Yang et~al.}]{madaan2024self}
Aman Madaan, Niket Tandon, Prakhar Gupta, Skyler Hallinan, Luyu Gao, Sarah Wiegreffe, Uri Alon, Nouha Dziri, Shrimai Prabhumoye, Yiming Yang, et~al. 2024.
\newblock Self-refine: Iterative refinement with self-feedback.
\newblock \emph{Advances in Neural Information Processing Systems}, 36.

\bibitem[{Marcus et~al.(1993)Marcus, Santorini, and Marcinkiewicz}]{marcus-etal-1993-building}
Mitchell~P. Marcus, Beatrice Santorini, and Mary~Ann Marcinkiewicz. 1993.
\newblock \href {https://aclanthology.org/J93-2004} {Building a large annotated corpus of {E}nglish: The {P}enn {T}reebank}.
\newblock \emph{Computational Linguistics}, 19(2):313--330.

\bibitem[{Meta(2024)}]{llama3}
Meta. 2024.
\newblock Introducing meta llama 3: The most capable openly available llm to date.

\bibitem[{Pan et~al.(2024)Pan, Saxon, Xu, Nathani, Wang, and Wang}]{pan-etal-2024-automatically}
Liangming Pan, Michael Saxon, Wenda Xu, Deepak Nathani, Xinyi Wang, and William~Yang Wang. 2024.
\newblock \href {https://doi.org/10.1162/tacl_a_00660} {Automatically correcting large language models: Surveying the landscape of diverse automated correction strategies}.
\newblock \emph{Transactions of the Association for Computational Linguistics}, 12:484--506.

\bibitem[{Raunak et~al.(2023)Raunak, Sharaf, Wang, Awadalla, and Menezes}]{raunak-etal-2023-leveraging}
Vikas Raunak, Amr Sharaf, Yiren Wang, Hany Awadalla, and Arul Menezes. 2023.
\newblock \href {https://doi.org/10.18653/v1/2023.findings-emnlp.804} {Leveraging {GPT}-4 for automatic translation post-editing}.
\newblock In \emph{Findings of the Association for Computational Linguistics: EMNLP 2023}, pages 12009--12024, Singapore. Association for Computational Linguistics.

\bibitem[{Shi et~al.(2023)Shi, Ajith, Xia, Huang, Liu, Blevins, Chen, and Zettlemoyer}]{shi2023detecting}
Weijia Shi, Anirudh Ajith, Mengzhou Xia, Yangsibo Huang, Daogao Liu, Terra Blevins, Danqi Chen, and Luke Zettlemoyer. 2023.
\newblock \href {https://arxiv.org/abs/2310.16789} {Detecting pretraining data from large language models}.
\newblock \emph{Preprint}, arXiv:2310.16789.

\bibitem[{Tian et~al.(2024)Tian, Xia, and Song}]{tian-etal-2024-large}
Yuanhe Tian, Fei Xia, and Yan Song. 2024.
\newblock \href {https://doi.org/10.18653/v1/2024.acl-long.384} {Large language models are no longer shallow parsers}.
\newblock In \emph{Proceedings of the 62nd Annual Meeting of the Association for Computational Linguistics (Volume 1: Long Papers)}, pages 7131--7142, Bangkok, Thailand. Association for Computational Linguistics.

\bibitem[{Tong et~al.(2024)Tong, Li, Wang, Wang, Teng, and Shang}]{tong-etal-2024-llms}
Yongqi Tong, Dawei Li, Sizhe Wang, Yujia Wang, Fei Teng, and Jingbo Shang. 2024.
\newblock \href {https://doi.org/10.18653/v1/2024.acl-long.169} {Can {LLM}s learn from previous mistakes? investigating {LLM}s{'} errors to boost for reasoning}.
\newblock In \emph{Proceedings of the 62nd Annual Meeting of the Association for Computational Linguistics (Volume 1: Long Papers)}, pages 3065--3080, Bangkok, Thailand. Association for Computational Linguistics.

\bibitem[{Tyen et~al.(2023)Tyen, Mansoor, Chen, Mak, and C{\u{a}}rbune}]{tyen2023llms}
Gladys Tyen, Hassan Mansoor, Peter Chen, Tony Mak, and Victor C{\u{a}}rbune. 2023.
\newblock Llms cannot find reasoning errors, but can correct them!
\newblock \emph{arXiv preprint arXiv:2311.08516}.

\bibitem[{Wang et~al.(2023)Wang, Wei, Schuurmans, Le, Chi, Narang, Chowdhery, and Zhou}]{wang2023selfconsistencyimproveschainthought}
Xuezhi Wang, Jason Wei, Dale Schuurmans, Quoc Le, Ed~Chi, Sharan Narang, Aakanksha Chowdhery, and Denny Zhou. 2023.
\newblock \href {https://arxiv.org/abs/2203.11171} {Self-consistency improves chain of thought reasoning in language models}.
\newblock \emph{Preprint}, arXiv:2203.11171.

\bibitem[{Wei et~al.(2022)Wei, Wang, Schuurmans, Bosma, ichter, Xia, Chi, Le, and Zhou}]{NEURIPS2022_9d560961}
Jason Wei, Xuezhi Wang, Dale Schuurmans, Maarten Bosma, brian ichter, Fei Xia, Ed~Chi, Quoc~V Le, and Denny Zhou. 2022.
\newblock \href {https://proceedings.neurips.cc/paper_files/paper/2022/file/9d5609613524ecf4f15af0f7b31abca4-Paper-Conference.pdf} {Chain-of-thought prompting elicits reasoning in large language models}.
\newblock In \emph{Advances in Neural Information Processing Systems}, volume~35, pages 24824--24837. Curran Associates, Inc.

\bibitem[{Xue et~al.(2005)Xue, Xia, Chiou, and Palmer}]{Xue2005ThePC}
Nianwen Xue, Fei Xia, Fu-Dong Chiou, and Martha Palmer. 2005.
\newblock \href {https://api.semanticscholar.org/CorpusID:9561000} {The penn chinese treebank: Phrase structure annotation of a large corpus}.
\newblock \emph{Natural Language Engineering}, 11:207 -- 238.

\bibitem[{Yang et~al.(2024)Yang, Yang, Hui, Zheng, Yu, Zhou, Li, Li, Liu, Huang, Dong, Wei, Lin, Tang, Wang, Yang, Tu, Zhang, Ma, Xu, Zhou, Bai, He, Lin, Dang, Lu, Chen, Yang, Li, Xue, Ni, Zhang, Wang, Peng, Men, Gao, Lin, Wang, Bai, Tan, Zhu, Li, Liu, Ge, Deng, Zhou, Ren, Zhang, Wei, Ren, Fan, Yao, Zhang, Wan, Chu, Liu, Cui, Zhang, and Fan}]{qwen2}
An~Yang, Baosong Yang, Binyuan Hui, Bo~Zheng, Bowen Yu, Chang Zhou, Chengpeng Li, Chengyuan Li, Dayiheng Liu, Fei Huang, Guanting Dong, Haoran Wei, Huan Lin, Jialong Tang, Jialin Wang, Jian Yang, Jianhong Tu, Jianwei Zhang, Jianxin Ma, Jin Xu, Jingren Zhou, Jinze Bai, Jinzheng He, Junyang Lin, Kai Dang, Keming Lu, Keqin Chen, Kexin Yang, Mei Li, Mingfeng Xue, Na~Ni, Pei Zhang, Peng Wang, Ru~Peng, Rui Men, Ruize Gao, Runji Lin, Shijie Wang, Shuai Bai, Sinan Tan, Tianhang Zhu, Tianhao Li, Tianyu Liu, Wenbin Ge, Xiaodong Deng, Xiaohuan Zhou, Xingzhang Ren, Xinyu Zhang, Xipin Wei, Xuancheng Ren, Yang Fan, Yang Yao, Yichang Zhang, Yu~Wan, Yunfei Chu, Yuqiong Liu, Zeyu Cui, Zhenru Zhang, and Zhihao Fan. 2024.
\newblock Qwen2 technical report.
\newblock \emph{arXiv preprint arXiv:2407.10671}.

\bibitem[{Yang et~al.(2022{\natexlab{a}})Yang, Tian, Peng, and Klein}]{yang-etal-2022-re3}
Kevin Yang, Yuandong Tian, Nanyun Peng, and Dan Klein. 2022{\natexlab{a}}.
\newblock \href {https://doi.org/10.18653/v1/2022.emnlp-main.296} {Re3: Generating longer stories with recursive reprompting and revision}.
\newblock In \emph{Proceedings of the 2022 Conference on Empirical Methods in Natural Language Processing}, pages 4393--4479, Abu Dhabi, United Arab Emirates. Association for Computational Linguistics.

\bibitem[{Yang et~al.(2022{\natexlab{b}})Yang, Cui, Ning, Wu, and Zhang}]{yang-etal-2022-challenges}
Sen Yang, Leyang Cui, Ruoxi Ning, Di~Wu, and Yue Zhang. 2022{\natexlab{b}}.
\newblock \href {https://doi.org/10.18653/v1/2022.findings-acl.11} {Challenges to open-domain constituency parsing}.
\newblock In \emph{Findings of the Association for Computational Linguistics: ACL 2022}, pages 112--127, Dublin, Ireland. Association for Computational Linguistics.

\bibitem[{Zhou et~al.(2023)Zhou, Hou, Li, Wang, Wang, Duan, and Zhang}]{zhou2023largelanguagemodelsunderstand}
Houquan Zhou, Yang Hou, Zhenghua Li, Xuebin Wang, Zhefeng Wang, Xinyu Duan, and Min Zhang. 2023.
\newblock \href {https://arxiv.org/abs/2311.08287} {How well do large language models understand syntax? an evaluation by asking natural language questions}.
\newblock \emph{Preprint}, arXiv:2311.08287.

\end{thebibliography}

\appendix
\newpage
\section{Appendix}

\subsection{Prompt for Constituency Parsing}\label{sec:baseline_prompt}
\begin{figure}[tb]
    \centering%
    \newtcolorbox{promptbox}[1]{%
        left=0pt,
        right=0pt,
        top=0pt,
        bottom=0pt,
        boxsep=6pt,
        colframe=black,
        title={#1},
    }
    \begin{minipage}[b]{\columnwidth}
        \begin{promptbox}{\textbf{LLM Prompt}}
            \scriptsize
            \textbf{You will be given one sentence for constituency parsing. Every word that is separated by a space should be considered an independent word and have its own constituency label. Please parse the sentence with given words.}\\
            Here are some examples:\\
            < sentence 1 >\enspace< parse tree 1 > \\
            < sentence 2 >\enspace< parse tree 2 > \\
            ...\\
            \textbf{Input: }Albany escaped embarrassingly unscathed .\\
            \textbf{Output: }(S (NP (NNP Albany)) (VP (VBD escaped) (S (ADJP (RB embarrassingly) (JJ unscathed)))) (. .))
        \end{promptbox}
    \end{minipage}
    \caption{The prompt for few-shot learning.}
    \label{fig:prompt_examples}
\end{figure}

Previous research \citep{bai2023constituencyparsingusingllms, tian-etal-2024-large} has demonstrated that linearizing constituency trees into sequences is effective when using LLMs for tree parsing. Therefore, following \citet{bai2023constituencyparsingusingllms}, we represent constituency tree structures using bracket notation, such as ``(S (NP (PRP He)) (VP (MD could) (RB not) (VP (VB speak))))''. We conduct experiments under the five-shot setting, randomly sampling five examples from existing annotated treebanks for in-context learning, as is shwon in Figure \ref{fig:prompt_examples}.

\subsection{Hint for Unmatch Correction}\label{sec:unmatch_correct}
\begin{figure}
    \centering
    \includegraphics[width=\columnwidth]{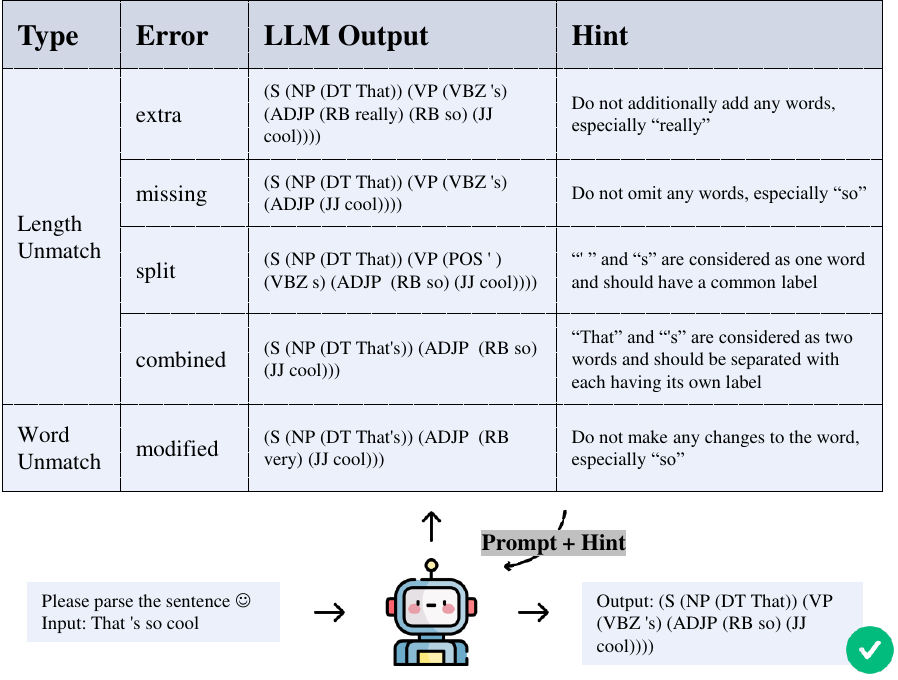}
    \caption{The process of Unmatch Correction.}
    \label{fig:hint}
\end{figure}

Following the evaluation script of EVALB, we categorize unmatch errors into two types: length unmatch and word unmatch. 
Length unmatch means the number of words in the predicted tree differ from the original sentence, while word unmatch indicates a word in the sentence has been altered.
As is shown in Figure \ref{fig:hint}, we design specific hints for different errors. 
By highlighting in the prompt the need to avoid mistakes from the previous answer, we guide the LLMs to generate a new answer.

\begin{table}[t]
    \centering
    \begin{tabular}{l cccc}
        \toprule        
        & \textbf{WikiMIA} & \textbf{PTB} &\textbf{ CTB} & \textbf{MCTB } \\
        \midrule    
        \textbf{MKP} &  7.13 &  9.59 &  8.95 &  9.53 \\ 
        \bottomrule
    \end{tabular}
    \caption{The MKP of different datasets with Llama-8B. A higher MKP suggests that the sentence is less likely to be included in the pre-training data of LLMs.}
    \label{tab:MKP}
\end{table}  
\subsection{The Examination of the Possibility of Contamination in Test Set}\label{sec:MKP}
{To examine the possibility of contamination in test set, we use the Min K\% probability (MKP) \cite{shi2023detecting}, as a metric to determine the extent to which the datasets are included in the pre-training data of LLMs. We follow the recommendation in \citet{shi2023detecting}, setting K value to 20\% and calculating the MKP of sentences in the three datasets under the Llama-8B model. We also report the MKP of the WikiMIA dataset for reference, which is used by \citet{shi2023detecting} to evaluate the identification capabilities for detecting pre-training data, consists of sentences from Wikipedia articles. Given the popularity of Wikipedia as a pre-training source, these sentences are highly likely to be part of the pre-training data for LLMs. 
As is shown in Table \ref{tab:MKP}, the MKP of all three constituency parsing datasets is significantly higher than that of WikiMIA. This suggests that these datasets are unlikely to have been included in the pre-training data of Llama-8B.
}

\begin{table}[t]
    \centering
    \begin{tabular}{l ccc}
        \toprule        
        & \textbf{R} & \textbf{P} &\textbf{F}  \\
        \midrule    
        3 rules*2=6 examples & 80.75 & 85.08 & 82.85\\ 
        5 rules*1=5 examples & \textbf{82.27} & \textbf{84.78} & \textbf{83.50}\\ 
        \bottomrule
    \end{tabular}
    \caption{The results of different numbers of the final examples.}
    \label{tab:setting}
\end{table}  
\subsection{The Effect of Different Numbers of the Final Rules and Examples}\label{sec:setting}
{To investigate the effect of different numbers of the final rules and examples, we add the experiment with the setting that select the top three rules with each rule providing two examples as shown in the first row of the Table \ref{tab:setting}. The second row shows the setting in our paper. Obviously, the original setting in our paper achieves better performance.}

\begin{table}[t]
    \centering
    \begin{tabular}{c ccc}
        \toprule        
        Method &   \textbf{R}   &   \textbf{P} &   \textbf{F}\\
        \midrule    
        POS searching  &  78.07  &  84.09  &  80.97 \\  
        Our method  & \textbf{ 82.27}  &  \textbf{84.78 } & \textbf{ 83.50} \\ 
        \bottomrule
    \end{tabular}
    \caption{Results of searching based on the POS-tag and our method on the PTB with GPT-4.}
    \label{tab:pos}
\end{table}  
\subsection{Different Searching Strategies}\label{sec:pos}
{We also compare our method with searching examples with the most similar POS sequences. As is shown in Table \ref{tab:pos}, our method based on the error-specific processing and ranking according to the label sequence significantly surpass ranking according to the POS. The POS-based ranking method only considers information from the lowest level leaf nodes, ignoring the structured information at intermediate levels. In contrast, our proposed method takes into account the structural information and can cover four types of errors, thus performing better.}

\begin{table}[t]
\centering
\small
\begin{tabular}{l|llll}
\toprule
\textbf{Model} & \textbf{Span}& \textbf{Label}& \textbf{Flatness} & \textbf{Deepness} \\
\midrule
Berkeley& \wz1,386&  \wz\wz944 & \wz2,736 & \wz2,985 \\
LLaMa-8B& 28,665&  \wz7,462 & \wz1,524 & \wz\wz964\\
LLaMa-70B & 12,404& \wz5,364 & \wz3,364 & \wz1,492\\
GPT-3.5  & \wz7,712 & \wz3,383 & \wz4,776 & \wz1,991\\
GPT-4  & \wz4,944 & \wz3,228 &  \wz3,791 & \wz 1,254\\
\bottomrule
\end{tabular}
\caption{The overall of four types of errors made by different models on the PTB.}
\label{table:ptb_error}
\vspace{-0.9em}
\end{table}

 
 

\begin{table}[t]
\centering
\small
\begin{tabular}{l|llll}
\toprule
\textbf{Model} & \textbf{Span}& \textbf{Label}& \textbf{Flatness} & \textbf{Deepness} \\
\midrule
Berkeley& 1,394&  1,397& 1,153 & 1,246 \\
LLaMa-8B& 2,830&  1,679 & \wz247 & \wz186\\
Qwen-72B & 1,987& 1,450 & \wz722 & \wz596\\
DeepSeek-v3 & 1,731& 1,110 & \wz740 & \wz487\\
GPT-3.5  & 1,752 & 1,046 & \wz797 & \wz514 \\
GPT-4  & 1,951 & 1,416 & \wz864 & \wz732   \\
\bottomrule
\end{tabular}
\caption{The overall of four types of errors made by different models on the CTB.}
\label{table:ctb_error}
\vspace{-0.9em}
\end{table}

 
 

\begin{table}[t]
\centering
\small
\begin{tabular}{l|llll}
\toprule
\textbf{Model} & \textbf{Span}& \textbf{Label}& \textbf{Flatness} & \textbf{Deepness} \\
\midrule
Berkeley& \wz\wz463 &  \wz\wz562 & \wz\wz600 & \wz\wz140 \\
LLaMa-8B& 11,609 &  \wz2,962 & \wz\wz493 & \wz\wz352 \\
LLaMa-70B & \wz4,704& \wz2,434 & \wz1,162 & \wz\wz449 \\
GPT-3.5  & \wz3,364 & \wz1,857 & \wz1,483 & \wz\wz718\\
GPT-4  & \wz1,570 &  \wz1,379 & \wz\wz939  & \wz\wz808 \\
\bottomrule
\end{tabular}
\caption{The overall of four types of errors made by different models on the MCTB.}
\label{table:mctb_error}
\vspace{-0.9em}
\end{table}

 
 

\subsection{Whether Our method Add New Errors?}\label{sec:add}
{we conducte a detailed analysis of the instances where originally correct parsing trees are corrected to be wrong during the process. Taking the parsing results of GPT-4 on the PTB as an example, initially, there are 24,203 grammar rules that were correct and consistent with the gold trees. After applying our method, 23,279 of these rules remain correct, indicating that our method has a wrong correction rate of only 4\%. Considering that the correction brings the improvement of over 10 F1 scores, it indicates that the correct corrections far outweigh the adding errors.}

\subsection{The Distribution of Four Types of Errors across Different models}\label{sec:all_errors}
{We provide the distributions of four types of errors across different models and different datasets in Table \ref{table:ptb_error}, Table \ref{table:ctb_error} and Table \ref{table:mctb_error}. Span error and label error are the most and second most frequent in the parsing results of different models. Moreover, as the performance of models declines, the proportion of span errors increases, indicating that models with weaker parsing capabilities find it more challenging to correctly segment spans.}

\end{CJK*}
\end{document}